\pdfoutput=1
\documentclass[11pt]{article}

\usepackage[final]{acl}

\usepackage{times}
\usepackage{latexsym}
\usepackage{amsmath}
\usepackage{amssymb}
\usepackage{multirow}
\usepackage[T1]{fontenc}
\usepackage[utf8]{inputenc}

\usepackage{microtype}

\usepackage{inconsolata}

\usepackage{graphicx}
\usepackage{listings}

\title{IPS: In-Prompt Process Supervision for Short Video Content Moderation}

\author{
  Mingchao Liu \quad Yu Sun \quad Ruixiao Sun \quad Xin Dong \\
  \textbf{Xiang Shen \quad Hongwei Wang \quad Hongyu Xiong \quad Yang Song} \\[0.3em]
  TikTok, Inc. \\
  \texttt{\{gorden.liu, yu.sun, ruixiao.sun, xindong, xiang.shen, hongwei.w, hongyu.xiong\}@tiktok.com} \\
  \texttt{ys@sonyis.me}
}

\begin{document}

\newcommand{\pheadNoSpace}[1] {\noindent\textbf{#1.}} % for initial headings
\newcommand{\pheadWithSpace}[1] {\vspace{1.25mm}\noindent\textbf{#1.}} % for subsequent 

\maketitle
% long version
% \begin{abstract}
% The advanced processing and reasoning capabilities of multimodal large language models (MLLMs) have driven substantial progress in vision-language (VL) understanding tasks. While MLLMs are effective at analyzing the overall semantics of content, they often lose focus on policy-mandated details required for content moderation. To address this limitation, we introduce \textit{IPS}, a novel framework that integrates In-Prompt Process Supervision into MLLMs by sequentially reasoning over ancillary questions during fine-tuning. 
% \textit{IPS} achieves significant improvements over baseline MLLMs on both public benchmarks and proprietary datasets. Notably, replacing human annotations with MLLM-generated ancillary labels results in only minimal performance degradation, demonstrating the method's robustness to noisy data and its scalability when leveraging MLLMs as annotators.
% These results establish \textit{IPS} as a scalable and effective solution for complex multimodal classification in large-scale industrial applications.
% \end{abstract}

% compact version
\begin{abstract}
Multimodal large language models (MLLMs) are effective at capturing the semantics of short video content; however, they often fail to attend to the policy-specific details required for reliable content moderation.
To address this limitation, we introduce \textit{IPS}, a novel framework that integrates In-prompt Process Supervision into MLLMs by introducing sequential reasoning over ancillary questions during fine-tuning. 
\textit{IPS} consistently outperforms baseline MLLMs on public and proprietary benchmarks.
Moreover, replacing human-annotated ancillary labels with MLLM-generated ones results in only marginal performance degradation, demonstrating robustness to noisy supervision and strong scalability with model-generated annotations.
These findings establish \textit{IPS} as a scalable and effective solution for complex multimodal classification in large-scale industrial settings.
\end{abstract}

\section{Introduction}
The rapid advancement of LLMs and MLLMs, such as GPT \cite{achiam2023gpt}, Gemini \cite{team2023gemini}, LLaVA \cite{li2024llava}, and Qwen \cite{wang2024qwen2, qwen2.5-VL}, has demonstrated remarkable capabilities across a wide range of applications, including visual question answering and contextual understanding. % These models excel at identifying fine-grained features in visual content and text while interpreting complex interactions across modalities. %, which supports their generalizability for diverse vision-language (VL) tasks.

Despite these advances, the effectiveness of MLLMs in highly specialized or sensitive domains remains underexplored.
Standard end-to-end supervised fine-tuning (SFT) often falls short of meeting the nuanced requirements of domain-specific tasks.
Unlike general content understanding tasks where MLLMs can leverage broad world knowledge, content moderation demands sophisticated reasoning grounded in complex and detail-intensive governance policies.
The intricacy of these policies, combined with the challenges of consistent interpretation, presents substantial difficulties even for human annotators.

% long version
% For example, determining whether a social media post is unoriginal requires nuanced evaluation. Crowd-sourced annotators follow predefined criteria by answering structured questions, such as "Does the content involve copyrighted material?" and "Are there user-generated edits?" A movie clip that would typically be considered unoriginal may instead be labeled as original if the content creator adds in-depth analysis or humorous elements. These tasks underscore the need for structured reasoning and fine-grained decision-making.

% compact version
For instance, determining whether a social media post constitutes unoriginal content requires nuanced judgment. Human annotators follow predefined criteria by answering structured questions (e.g., whether the content contains copyrighted material or meaningful user-generated edits).
In this way, a movie clip that would typically be labeled unoriginal may instead be considered original if the creator adds substantive commentary or humor.
Such scenarios underscore the need for structured reasoning and fine-grained decision-making. \cite{lan2025contextual}

\begin{figure*}[htbp]
    \centering
    \includegraphics[width=0.7 \linewidth]{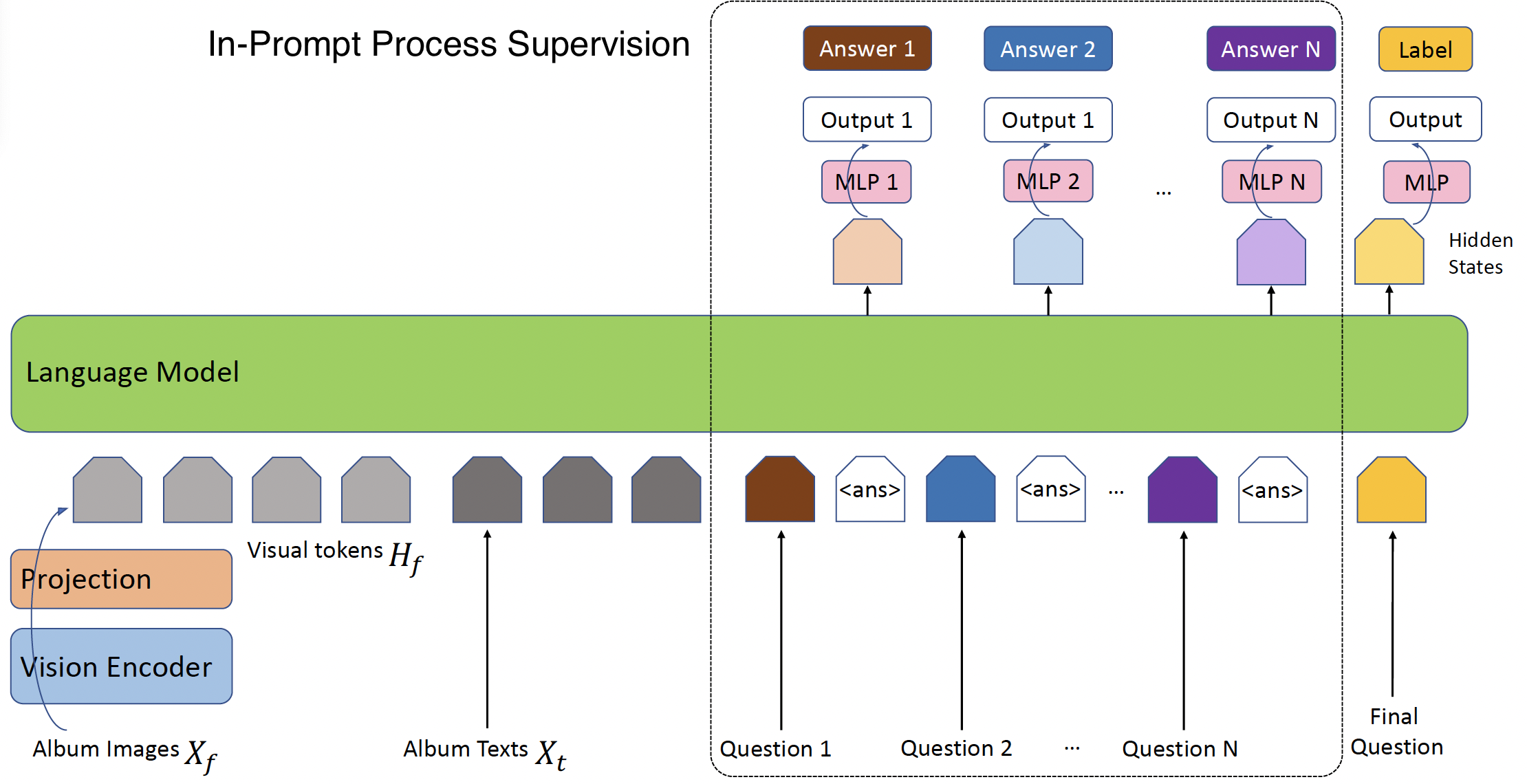} % Adjust the width as needed
    \caption{In-prompt Process Supervision (\textit{IPS}) framework. The framework comprises a vision encoder, a modality-alignment projector, and a language model that jointly processes visual and textual tokens. During SFT, $N$ questions are concatenated into a single prompt to improve both training and deployment efficiency. Each ancillary question terminates with an \texttt{<ans>} token; the hidden state corresponding to this token is passed through an MLP to predict the associated answer.}
    %\caption{In-Prompt Process Supervision(\textit{IPS}) Framework. It comprises three key components: a vision encoder for handling images, a vision-language modality alignment projector, and a language model that integrates visual and textual tokens. The framework integrates $N+1$ questions into one input prompts during the SFT process for efficiency in training and future deployment. Each ancillary question concludes with an \texttt{<ans>} token, whose hidden state output is processed by a multi-layer perceptron (MLP) to predict the answer for the corresponding ancillary question. }
    \label{fig:sequential_qa_design}
\end{figure*}

In this paper, we propose a novel and effective approach to enhance the domain-specific supervised fine-tuning of MLLMs for complex content moderation tasks.
Our work introduces two key innovations:

\textbf{\textit{IPS} framework}.
We present \textit{IPS}, a structured process supervision framework that aligns MLLM reasoning with human annotation workflows. \textit{IPS} leverages ancillary labels to sequentially guide the model's decision-making. Specifically, each training instance embeds multiple question–answer pairs in sequence, where \texttt{<ans>} tokens mark supervision points and the corresponding answers serve as intermediate (ancillary) supervision signals.

We evaluate \textit{IPS} on a public Hate Speech Detection (HSD) dataset from MM-Soc \cite{jin2024mmsocbenchmarkingmultimodallarge}, as well as two proprietary tasks: Unoriginal Content Classification (UCC) and Adult Nudity and Sexual Activity (ANSA) detection. Experimental results demonstrate substantial improvements in F1 score and recall across multiple precision thresholds compared to baseline MLLMs trained with standard end-to-end SFT. Ablation studies further confirm that both the inclusion of ancillary labels and the structured process supervision design are critical to these gains.

\textbf{Scaling process supervision with MLLM-generated annotations}.
While process supervision improves performance, it introduces additional annotation overhead beyond final-label supervision. Notably, ancillary labels serve as intermediate guidance rather than strict determinants of the final outcome, making them less sensitive to minor inaccuracies. Motivated by this observation, we explore automatically generating ancillary labels using general-purpose MLLMs.

Despite achieving only 75\% agreement with human annotations, \textit{IPS} remains robust to such noise. On the UCC task, replacing human-annotated process labels with MLLM-generated ones results in only marginal performance changes, demonstrating the scalability of our approach for large-scale industrial deployment.

\section{Related Work}

\subsection{Multimodal Large Language Models (MLLMs)}
Recent advancements in Multimodal Large Language Models (MLLMs), such as GPT-4V~\cite{achiam2023gpt}, GPT-4o~\cite{achiam2023gpt}, Gemini~\cite{team2023gemini}, and Claude-3.5, demonstrate remarkable versatility across a wide range of vision tasks, including single-image, multi-image, and video analysis. 
Although many models are tailored for specific task types, the open-source model LLaVA-OneVision~\cite{li2024llava} is designed for strong cross-task performance and exhibits robust feature transfer.
Several other versatile models, such as Video-LLaMA~\cite{zhang2023video}, %which integrates visual and auditory content understanding, 
and VILA~\cite{lin2024vila}, 
further highlight the growing potential of multi-scenario MLLMs. However, applying MLLMs in classification tasks remains challenging, as critical information for classification is not efficiently extracted by LLMs~\cite{VLMClassifier, li2024comae,10.1145/3711896.3737195, li2025lion, meng2026tri, chen2026omnivideo}. 

\subsection{Chain of Thought} 
Chain-of-Thought (CoT) prompting, introduced by \citet{wei2022chain}, has emerged as a powerful method to enable process supervision in LLMs. 
Leveraging structured prompting, CoT decomposes complex tasks into intermediate reasoning steps, allowing models to solve problems systematically. 
Building upon this foundation, \citet{yao2024tree} and \citet{long2023large} introduce a Tree-of-Thought (ToT) framework, a novel approach to multi-round question answering (QA) using language models. 
\citet{besta2024graph} propose Graph-of-Thought (GoT) to model the reasoning process as a graph. 
In addition, fine-tuning with CoT \cite{yuan2023scaling, xiang2025promptsculptormultiagentbasedtexttoimage, wang-etal-2025-reasoning-enhanced} demonstrates better performance compared with direct SFT in reasoning tasks. Collectively, these studies highlight that leveraging contextual influence during the reasoning process can substantially enhance the performance of LLMs.

\subsection{Process Supervision} \citet{uesato2022solving} introduces a comparison between process supervision and outcome supervision in reasoning tasks and \citet{lightman2023let, li2025preference, zhou2026comem} follow the study and shows that process-supervised reward models (PRM) lead to a significant improvement in mathematical reasoning. 
%The new approaches achieved success in various fields after mathematical reasoning. 
\citet{ma2023adapting},  transforms content moderation into a reasoning task and provides weak supervision in fine-tuning LLMs. \citet{wang2024process} applies the PRMs to verify the generation of clinical notes.

\section{Methodology}

\subsection{Problem Formulation}
Given a video $X$ and a set of violation labels $L$ defined by a predefined policy, the content moderation model aims to identify the label in $L$ that best characterizes the content, along with a confidence score. This score is subsequently used by the recommendation system to determine whether the content should be recommended.

\subsection{Dataset Structure}
Each policy is associated with a dedicated dataset annotated by professionally trained human reviewers in accordance with the corresponding policy documentation.
In industrial content-moderation pipelines, such policies are typically operationalized as structured decision trees so that annotators can reach consistent judgments on complex cases.
For each content instance $X$, the annotator follows a predefined decision tree $T$, producing a final label $L_f$ along with a set of ancillary labels $(L_1,..., L_m)$ corresponding to the intermediate decision nodes.

\begin{figure}[ht]
    \centering  
    \includegraphics[width=5cm]{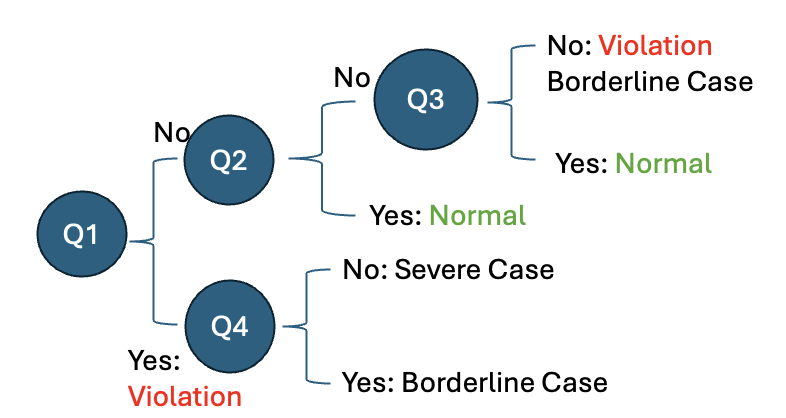}
    \caption{Illustration of the decision tree used by labeling agents. In practice, annotators may not strictly adhere to the tree in order to maintain flexibility, and only the final Normal/Guilty label undergoes formal quality assurance.}
    \label{fig:decision-tree}
\end{figure}

For each data entry $(X, Y)$, the input is defined as $X = (X_f, X_t)$, where $X_f$ denotes the visual frames and $X_t$ represents the textual component of the content. The corresponding label is $Y = [L_f, [L_1, \ldots, L_m]]$. Here $L_f$ is the final label to be predicted (typically binary), while $L_1, \ldots, L_m$ are ancillary labels derived from the intermediate nodes of the decision tree $T$ (e.g., Figure~\ref{fig:decision-tree}). These ancillary labels either function as intermediate indicators that guide annotators toward the final decision or serve as fine-grained sub-labels (e.g., violation sub-types) that enrich the annotation. The number of ancillary labels $m$ may vary even within the same policy, as certain branches of the decision tree can terminate early and some questions are optional.

Due to resource constraints, only the final label $L_f$ undergoes formal quality assurance, while ancillary labels remain relatively noisy and are therefore often ignored in prior work. This underutilization represents a significant loss of valuable annotation signals. To address this issue, we propose \textit{IPS}, a cost-efficient framework that effectively leverages these noisy ancillary labels to better align the model with policy requirements and improve final-label prediction performance.

\subsection{\textit{IPS}: In-Prompt Process Supervision}
% long version
% As illustrated in \textbf{Figure~\ref{fig:sequential_qa_design}}, we introduce In-Prompt Process Supervision (\textit{IPS}), a novel framework that incorporates multiple question prompts into the input and uses their corresponding answers as process labels to provide process supervision during MLLM fine-tuning. Intuitively, this design resembles asking a series of simpler, related questions before posing a difficult one, simulating multi-round guidance within a single inference pass. This design introduces negligible inference-time overhead, leverages noisy ancillary labels that were previously unused, and effectively enhances final-question prediction accuracy.

% compact version
As illustrated in Figure~\ref{fig:sequential_qa_design}, \textit{IPS} incorporates multiple question prompts and their corresponding answers as process-level supervision signals during MLLM fine-tuning. Analogous to addressing simpler, related questions before tackling a complex one, this design simulates multi-step guidance within a single forward pass. It introduces negligible inference overhead, leverages previously unused ancillary labels, and improves the accuracy of the final prediction.

% \textbf{Vision Encoder \& VL Modality Alignment Projector.}
\paragraph{Model Architecture}
For each video frame or album image, the visual input $X_f$ is first processed by a vision encoder $g(\cdot; \theta_g)$ to extract visual features:
\begin{equation}
    Z_f = g(X_f; \theta_g).
\end{equation}
The extracted features are then passed through a projector module $p(\cdot; \theta_p)$, which maps them into a shared multimodal representation space:
\begin{equation}
    H_f = p(Z_f; \theta_p).
\end{equation}
Here $Z_f$ captures high-level visual representations, while $H_f$ aligns them with the textual modality for joint modeling.

% \textbf{Language Model for Visual and Text Tokens.}
The language model $LM(\cdot; \theta_{\text{lm}})$ processes a multimodal token sequence of length \( L \), consisting of visual tokens \( H_f \), textual tokens \( X_t \), and a policy-specific prompt $P$.
The model typically adopts a decoder-only transformer architecture, which produces a sequence of hidden representations $H$ corresponding to each input token:
\begin{equation}
    H = LM([H_f,X_t,P]; \theta_{\text{lm}}).
\end{equation}

\paragraph{\textit{IPS} and Classification Layers}
During SFT, $N$ rounds of ancillary question-answer pairs are incorporated into the prompt $P$ to provide process supervision:
\begin{equation}
\begin{aligned}
P =\;& ``\{question1\}\texttt{<ans>}\{question2\}\texttt{<ans>} \\
    & \ldots \{final\_question\}''
\end{aligned}
\end{equation}

Each ancillary question $i$ is followed by a special \texttt{<ans>} token. Let $d$ denote the hidden-state dimension of the language model. The hidden state corresponding to this token, $H_{<ans>}^{i} \in \mathbb{R}^{d}$, is fed into an MLP classifier $f_\mathrm{cl}^{i}(\cdot; \theta_\mathrm{cl}^{i})$, which projects it into the answer space:
\begin{equation}
    \hat{y}^{i} = f_\mathrm{cl}^{i}(H_{<ans>}^{i}; \theta_\mathrm{cl}^{i}),
\end{equation}
where $\hat{y}^{i} \in \mathbb{R}^{|\mathcal{Y}^{i}|}$ is the predicted distribution and $\mathcal{Y}^{i}$ denotes the set of possible classes for question $i$.
The hidden representation of the final token in the sequence is passed through another MLP to produce the prediction for the final label.

During training, the model minimizes a weighted sum of the losses from all $N$ ancillary questions and the final question:
\begin{equation}
\begin{aligned}
    &\text{Training\_Loss} = \sum_{i=1}^{N+1} w_i \cdot \mathcal{L}(\hat{y}^{i}, y^{i}).
\end{aligned}
\end{equation}

The weights $w_i$ are hyperparameters, with the default configuration:

\begin{align}
     w_i = 
    \begin{cases}
      0.1, & i = 1,\dots,N \ \text{(ancillary questions)} \\
      1, & i = N+1 \ \text{(final question)}
    \end{cases}
\end{align}

Here, $\mathcal{L}(\cdot)$ denotes the loss function (e.g., cross-entropy). We experimented with ancillary weights in $\{0, 0.05, 0.1, 0.15, 0.2, 1\}$ and found that a stable default around 0.1 performs robustly across multiple tasks when $N<5$. The fact that a single uniform weight works well across three heterogeneous tasks (UCC, ANSA, HSD) without per-task tuning indicates that the benefit of \textit{IPS} comes from its structural design rather than careful hyperparameter calibration. In settings where certain ancillary criteria function as hard vetoes in the policy (e.g., strict keyword or entity gates), a hierarchy-aware or learnable weighting scheme could be a natural extension, which we leave to future work.

The core intuition behind \textit{IPS} is twofold:

(1) \textit{Additional Task-Specific Information}.
Incorporating domain-tailored process labels provides richer supervision signals during SFT, enabling the model to better capture policy-relevant features.

% long version
% (2) \textit{Chain-of-Thought (CoT) Reasoning in Decoder-Only Architectures}: In \textit{IPS}, each ancillary question is paired with its corresponding representation at prior position. Placing the final question last allows the model to aggregate all prior process supervision signals before predicting the final label. This setup aligns with the Chain-of-Thought (CoT) reasoning paradigm. By supervising earlier parts of the sequence, the model’s hidden states of early positions evolve to reflect useful intermediate semantics, which ultimately improves reasoning coherence and boosts final prediction accuracy. The decoder-only architecture naturally facilitates this process, as a token’s hidden state attends to all prior tokens, enabling process supervision to influence the final decision.

% compact version
(2) \textit{Chain-of-Thought Reasoning in Decoder-Only Architectures}.
By placing ancillary questions before the final question, the model accumulates intermediate supervision signals prior to making the final prediction. This structure aligns naturally with the Chain-of-Thought paradigm. Supervising earlier tokens encourages hidden states to encode meaningful intermediate semantics, improving reasoning coherence and final prediction accuracy. The decoder-only architecture facilitates this process, as each token attends to all preceding tokens, allowing process supervision to directly influence the final decision.

\subsection{Process Annotation through MLLMs}
We leverage general-purpose MLLMs, such as GPT-4o and LLaVA, to perform zero-shot process labeling. Because each ancillary question is intentionally designed to be clear and narrowly scoped, the risk of hallucination is substantially lower than when directly predicting final labels. Moreover, we empirically demonstrate that \textit{IPS} remains robust under noisy process supervision. A detailed analysis of this robustness is provided in Section~\ref{analysis}.
This strategy enables \textit{IPS} to scale seamlessly within the supervised fine-tuning (SFT) pipeline of MLLMs for large-scale deployment.

\section{Experiments and Results}

We conducted our experiments on two proprietary datasets, UCC (Unoriginal Content Classification), ANSA (Adult Nudity and Sexual Activity), and one open-source dataset, MM-Soc HSD (Hate Speech Detection benchmark for Multimodal Large Language Models in social media platforms)~\cite{jin2024mmsocbenchmarkingmultimodallarge}. Across all evaluated datasets, models incorporating the proposed \textit{IPS} mechanism consistently outperform both baseline models and current state-of-the-art approaches, achieving top rankings on the leaderboard.

\subsection{Experiment Setup}

\subsubsection{Datasets}
% long version
% UCC task involves classifying posts as either OC (Original Content), where content is likely created by the publisher, or UC (Unoriginal Content), where content is likely copied from other sources. The experimental dataset comprises $150K$ samples for training and $8K$ for test. Each sample includes one or more images and accompanying text such as title. Each sample is human-annotated with a binary final label (OC or UC), and four binary sub-issue labels, which naturally serves as process supervision for this task.

% compact version
The UCC task classifies posts as OC (Original Content) or UC (Unoriginal Content). The dataset includes $150K$ training and $8K$ test samples, each containing images with accompanying text. Every sample has a binary final label (OC/UC) and four binary sub-issue labels that serve as process supervision.

% long version
% ANSA dataset is a large in-house dataset, which consists of 4 million training samples and 21K test samples and it is employed for pornographic content detection. Originally, this dataset was annotated with a single binary label indicating whether the content contains adult material, nudity, or sexual activity, including both explicit and implicit instances. It comes without any associated process annotations. Consequently, all process labels discussed in the ANSA task section are generated by MLLM. We employ the pre-trained LLaVA-7b model as the annotator for this in-house dataset.

% compact version
The ANSA dataset is a large in-house benchmark for pornographic content detection, comprising 4M training and 21K test samples. It was originally annotated with a single binary label indicating the presence of adult material, nudity, or sexual activity (explicit or implicit), without process annotations. Therefore, all process labels in the ANSA experiments are generated by MLLM, using a pre-trained LLaVA-7b model as the annotator.

Furthermore, we evaluate the \textit{IPS} method on the open-source HSD dataset from MM-Soc~\cite{jin2024mmsocbenchmarkingmultimodallarge}, which focuses on hateful meme detection. The dataset contains 8.5K training samples and 500 test samples. The three process labels used in this open-source dataset are generated with GPT-4o.

To show that \textit{IPS} can work together with traditional CoT-incorporated training~\cite{ma2023adapting} and keep up with the latest MLLM models, we conduct experiments with the Qwen2.5-VL-7B~\cite{qwen2.5-VL} model on the ANSA-borderline dataset, a subset of ANSA that introduces an even harder task of recognizing borderline content. This dataset consists of 200K training samples and 20K test samples. The CoT reasoning process in this dataset is generated with GPT-4o and the process supervision labels are provided by human labelers.

Details on model training are provided in Appendix~\ref{model_training}, while the specific prompts used for annotation can be found in Appendix~\ref{llm_prompt}~\&~\ref{llm_prompt_hate}.

\subsubsection{Metrics}
To assess model performance, we primarily evaluate the recall at various precision levels, alongside the F1 score on both public benchmark and industrial datasets. 

In addition, we deploy the best ANSA model online to build a content-based recommendation strategy and evaluate \textit{IPS} via an A/B test using Sexual Suggestive View Rate (SSVR) and Inappropriate Content View Rate (ICVR)—the percentage of traffic viewing sexual or inappropriate content.

% no need to add gbdt sinve we dont introduc what gbdt is for. 
\subsection{Key Results and Analyses}

\begin{table}[htbp]
\small
\space
\centering
\resizebox{\linewidth}{!}{\begin{tabular}{l | l | c  c  c}
    \hline
    \hline
     Task    &   Models &      F1    &     R@P60    &    R@P65  \\ 
    \hline
   \multirow{5}{*}{UCC}  & SigLIP~\cite{zhai2023sigmoid} &  - & 70.6 & 62.0  \\%\hline
    & LLaVA-OV-0.5B Vanilla &   66.7 & 71.6 & 67.5  \\%\hline
   &   LLaVA-OV-0.5B \textit{IPS}  & \textbf{68.9} & \textbf{76.4} & \textbf{73.2}  \\%\hline
   &   LLaVA-OV-7B Vanilla  & 68.9 & 75.4 & 72.2 \\%\hline
    &  LLaVA-OV-7B \textit{IPS} & \textbf{69.4} & \textbf{76.6} & \textbf{72.6} \\%\hline
    \hline
     \multirow{5}{*}{ANSA}  & X-VLM~\cite{bao2022vlmo} &  45.7 & 38.0 & 32.4  \\%\hline
   &    LLaVA-OV-0.5B Vanilla  & 54.9 & 49.9 & 44.1  \\%\hline
   &   LLaVA-OV-0.5B \textit{IPS}  & \textbf{56.4} & \textbf{52.8} & \textbf{46.2}  \\%\hline
   &   LLaVA-OV-7B Vanilla  & 56.6 & 52.0 & 46.6 \\%\hline
   &   LLaVA-OV-7B \textit{IPS} & \textbf{57.7} & \textbf{53.1} & \textbf{48.0} \\%\hline
    \hline
   \multirow{6}{*}{HSD} &   LLaVA-1.5-7B  & 49.0 &  - & -  \\%\hline
  &   LLaVA-1.5-13B    & 57.8 &  - & -  \\%\hline
   &    LLaVA-OV-0.5B Vanilla & 68.6 &  58.0 & 56.6  \\%\hline
  &    LLaVA-OV-0.5B \textit{IPS}  & \textbf{68.9} & \textbf{60.4} & \textbf{59.2}  \\%\hline
 &     LLaVA-OV-7B Vanilla  & 70.9 & 64.3 & 63.8 \\%\hline
  &   LLaVA-OV-7B \textit{IPS}  & \textbf{72.6} & \textbf{69.1} & \textbf{67.6} \\%\hline
    \hline
    \hline
    \end{tabular}
}
\caption{Overall Performance (in \%) comparison of Vanilla and \textit{IPS} on all experiment datasets.}
\label{tab:overall_performance}
\end{table}
\textbf{Table~\ref{tab:overall_performance}} shows that introducing the \textit{IPS} mechanism improves performance across all three datasets evaluated in our experiments consistently. 
This indicates that the \textit{IPS} method can be widely adopted for MLLM-based visual-language data classification tasks.

%%% UCC
In the UCC task, \textit{IPS} significantly improves offline recall over both the vanilla setting and the internal baseline SigLIP~\cite{zhai2023sigmoid} across various precision thresholds. These results highlight the effectiveness of \textit{IPS} in enhancing model reasoning by aligning it with human-defined logic, leading to stronger performance across different model scales.

%%% ANSA

The ANSA dataset is the largest and most challenging in-house dataset, featuring a continuously updated test set that adapts to emerging trends.
Despite the challenges posed by ANSA, the \textit{IPS} model demonstrates substantial improvements over the X-VLM-based online model, outperforming all other in-house models and achieving the highest ranking on the internal leaderboard.

%%% Hateful

On the HSD dataset from MM-Soc~\cite{jin2024mmsocbenchmarkingmultimodallarge}, \textit{IPS} again outperforms vanilla models in both F1 and precision-recall. Case studies further illustrate the interpretability and robustness of its sub-task predictions. See Appendix~\ref{case_study} for details.

\begin{table*}[htbp]
\small
\centering
\begin{tabular}{l | c | c | c  c  c  c  c}
\hline
\hline
 Models & Data Size & F1 &    R@P60    & R@P65 &     R@P70    & R@P75 &    R@P80  \\ 
\hline
LLaVA-OV-0.5b Vanilla & & 51.7 & 35.1 & 29.7 & 23.5 & 18.3 & 13.7  \\%\hline
LLaVA-OV-0.5b \textit{IPS} (w/ MLLM data) & 12k & \textbf{56.1} & \textbf{48.6} & \textbf{39.5} & \textbf{32.0} & \textbf{24.8} & \textbf{18.1} \\%\hline
LLaVA-OV-0.5b \textit{IPS} (w/ Human data) & & 55.5 & 45.3 & 36.1 & 29.8 & 20.2 & 15.6 \\%\hline
\hline
LLaVA-OV-0.5b Vanilla & & 56.2 & 50.8 & 46.1 & 41.4 & 31.7 & 24.3  \\%\hline
LLaVA-OV-0.5b \textit{IPS} (w/ MLLM data) & 24k & 59.5 & 58.4 & 52.4 & 46.1 & 38.7 & 31.2 \\%\hline
LLaVA-OV-0.5b \textit{IPS} (w/ Human data) & & \textbf{60.6} & \textbf{60.7} & \textbf{55.7} & \textbf{49.0} & \textbf{43.2} & \textbf{37.5} \\%\hline
\hline
LLaVA-OV-0.5b Vanilla & & 62.1 & 64.0 & 56.6 & 52.2 & 44.5 & 37.6  \\%\hline
LLaVA-OV-0.5b \textit{IPS} (w/ MLLM data) & 54k & \textbf{64.7} & \textbf{68.6} & \textbf{64.1} & 55.4 & 48.1 & 43.6 \\%\hline
LLaVA-OV-0.5b \textit{IPS} (w/ Human data) & & 64.4 & 66.5 & 63.0 & \textbf{59.1} & \textbf{53.2} & \textbf{48.0} \\%\hline
\hline
LLaVA-OV-0.5b Vanilla & & 65.7 & 70.5 & 65.9 & 58.4 & 52.4 & 46.5  \\%\hline
LLaVA-OV-0.5b \textit{IPS} (w/ MLLM data) & 90k & \textbf{66.8} & \textbf{72.7} & 67.1 & \textbf{62.9} & 57.7 & 49.6 \\%\hline
LLaVA-OV-0.5b \textit{IPS} (w/ Human data) & & 66.3 & 72.1 & \textbf{67.6} & 62.6 & \textbf{57.9} & \textbf{51.2} \\%\hline
\hline
\hline
\end{tabular}
\caption{Impact of using MLLM-generated process label to replace human annotated label on UCC dataset. Different sizes of training data are compared to demonstrate the consistency of this approach. (Performance displayed in \%)}
\label{tab:gpt_data}
\end{table*}

\section{Consistency Across Different Dataset Sizes}
% To evaluate the stability of \textit{IPS}, we trained models on progressively larger subsets of 854K, 1.7M, 2.5M, and 4M samples in the ANSA task, examining whether the impact of \textit{IPS} remains consistent across different dataset sizes.
% As shown in \textbf{Table~\ref{tab:ANSA}}, \textit{IPS} consistently outperforms its vanilla counterpart at all scales. This consistent margin across scales suggests that the introduction of \textit{IPS} remains beneficial as data availability varies.
% Notably, when trained on the full dataset, our model achieves a precision of 63 at recall 50, setting a new state-of-the-art on the internal leaderboard.
% Performance gains are most pronounced when training data are limited—a trend also observed in the UCC task, shown in \textbf{Table~\ref{tab:gpt_data}}. 

To assess the stability of \textit{IPS}, we trained models on progressively larger subsets (854K, 1.7M, 2.5M, and 4M samples) in the ANSA task. As shown in \textbf{Table~\ref{tab:ANSA}}, \textit{IPS} consistently outperforms the vanilla model across all scales, indicating stable benefits under varying data availability. With the full dataset, our model achieves 63 precision at 50 recall, establishing a new state-of-the-art on the internal leaderboard. Gains are most pronounced in low-data settings, a trend also observed in the UCC task (\textbf{Table~\ref{tab:gpt_data}})

\subsection{Feasibility of Replacing Human-Annotated Process Labels with LLM-Generated Labels}

To examine the feasibility of substituting human-provided process labels with those generated by MLLMs, we employed a general-purpose MLLM to produce labels for the UCC task, which already includes human annotations. We then trained models on subsets of varying sizes and compared the performance of \textit{IPS} using MLLM-generated labels against models trained with human-labeled data.

As shown in \textbf{Table~\ref{tab:gpt_data}}, \textit{IPS} trained on MLLM-generated data performs comparably to its human-labeled counterpart and consistently outperforms vanilla SFT across all data scales. These results demonstrate the robustness of \textit{IPS}, even with imperfect supervision, and confirm its ability to leverage noisy but structured intermediate signals.

Overall, this highlights the potential of MLLMs as a reliable and scalable process supervisors within the \textit{IPS} framework, offering a cost-effective alternative to manual annotation while maintaining strong performance in real-world settings.

\paragraph{Analysis of MLLM-generated Process Labels} \label{analysis}
Although \textit{IPS} is robust to imperfect process labels, assessing the quality of MLLM-generated annotations remains important. We manually reviewed 1,000 UCC samples, examining both process and final-label predictions to evaluate zero-shot process supervision.

As shown in \textbf{Table~\ref{tab:quality_assurance}}, zero-shot predictions for the final question achieved an overall accuracy of 57.6\%, with a missing rate of 12.23\% due to ambiguous questions. This lower performance is expected, as general-purpose MLLMs lack domain-specific knowledge. In contrast, process questions—being simpler—showed no missing values and much higher human–machine agreement. 
% We report a ‘human–machine agreement rate’ rather than absolute accuracy, since human-provided process labels lack formal quality assurance; only the final question’s consistency can be meaningfully evaluated. 
These results suggest that while general-purpose MLLMs cannot yet match human annotators for final-label generation, they are suitable for generating process supervision labels that require less stringent quality verification. Prompt examples are provided in Appendix~\ref{llm_prompt}.

\begin{table}[htbp]
\small
\space
\centering
\resizebox{\linewidth}{!}{\begin{tabular}{l | c }
    \hline
    \hline
     Question &    Agreement Rate    \\ 
    \hline
    Final Question (Is the content UCC) &   57.60  \\%\hline
    \hline
    Process Question 1 (Watermark)&  79.10 \\%\hline
    Process Question 2 (UGC)&  66.95  \\%\hline
    Process Question 3 (Text originality)&  74.29 \\%\hline
    % process question 4 (Title originality)&  51.13 \\%\hline
    Process Question 4 (Make a good serie)& 77.68 \\%\hline
    \hline
    \hline
    \end{tabular}
}
\caption{Zero-Shot Consistency with human-provided label (in \%) of MLLM annotator on UCC task. We report a ‘human–machine agreement rate’ rather than absolute accuracy, since human-provided process labels lack formal quality assurance; only the final question’s consistency can be meaningfully evaluated.} % We define ‘Agreement Rate’ as the proportion of cases where MLLM-generated process labels match human annotations.} % Since process annotations lack formal quality assurance, this metric primarily assesses consistency rather than absolute correctness.}
\label{tab:quality_assurance}
\end{table}

\section{Deployment Results}
\paragraph{Deploy \textit{IPS} online for ANSA governance}
We conducted A/B experiments over a 14-day period, involving 10\% of total traffic, yielding highly significant results (p < 0.001). \textit{ANSA-IPS} is compared against a legacy X-VLM\cite{bao2022vlmo} based model. We evaluated the online metrics of inappropriate content view rate (ICVR) and sexual suggestive view rate (SSVR). Following the deployment of the upgraded model, ICVR decreased by 0.22\% (95\% CI: 0.22\% ± 0.05\%) and SSVR decreased by 1.3\% (95\% CI: 1.3\% ± 0.08\%). These metrics are critical for online model deployment, indicating that the \textit{ANSA-IPS} model has an enhanced ability to reduce undesirable user experiences.

% long version
%\section{Conclusion}
%In this work, we present \textit{IPS}, a novel framework for tackling challenging multimodal content moderation tasks. By integrating ancillary labels through a sequentially structured process supervision mechanism, \textit{IPS} better aligns model reasoning with human labelers’ chain-of-thought reasoning (\textit{CoT}), achieving notable performance improvements over vanilla end-to-end SFT approaches. We further demonstrate the scalability of \textit{IPS} by substituting human-annotated ancillary labels with MLLM-generated ones, with only minimal degradation in performance. This approach significantly reduces the need for extra manual labeling, making it applicable to a broad spectrum of real-world tasks.

% compact version
\section{Conclusion}
In this work, we present \textit{IPS}, a novel framework for tackling challenging multimodal content moderation tasks. 
By integrating ancillary labels through a sequentially structured process supervision mechanism, \textit{IPS} better aligns model reasoning with human labelers’ chain-of-thought reasoning (\textit{CoT}), achieving notable performance improvements over vanilla end-to-end SFT approaches.  
Replacing human-annotated ancillary labels with MLLM-generated ones yields only minor performance drops, demonstrating scalability while reducing manual labeling effort for real-world applications.

% \begin{acks}
% To TikTok, for the data and the computing resources.
% \end{acks}

\section*{Ethical Considerations}
Content moderation systems inherently involve trade-offs between over-moderation, which unfairly restricts legitimate expression, and under-moderation, which exposes users to harmful content.
\textit{IPS} is designed to be mindful of both ends of this trade-off.
First, the live A/B evaluation in Section 6 shows that deploying \textit{IPS} reduces the Inappropriate Content View Rate (ICVR) by 0.22\% and the Sexual Suggestive View Rate (SSVR) by 1.3\% on real user traffic, indicating that the method reduces harmful exposure without indiscriminate over-flagging.
Second, beyond aggregate metrics, the interpretable ancillary predictions produced by \textit{IPS} serve as an operational safeguard: instead of treating the final label as a black box, human moderators can audit the model's intermediate judgments on borderline cases and override the final decision when the ancillary signals disagree with policy-specific nuance.
Finally, all proprietary training and evaluation data used in this work come from human-reviewed content-moderation pipelines operating under platform policies; no additional user data was collected for this study. The public HSD benchmark we evaluate on is released by \citet{jin2024mmsocbenchmarkingmultimodallarge} under its original license, and we use it strictly within the terms of that release.


\bibliography{acl_latex}

\appendix

\section{Ablation on \textit{CoT} reasoning structure}

\begin{figure}[ht]
    \centering  
    \includegraphics[width=7cm]{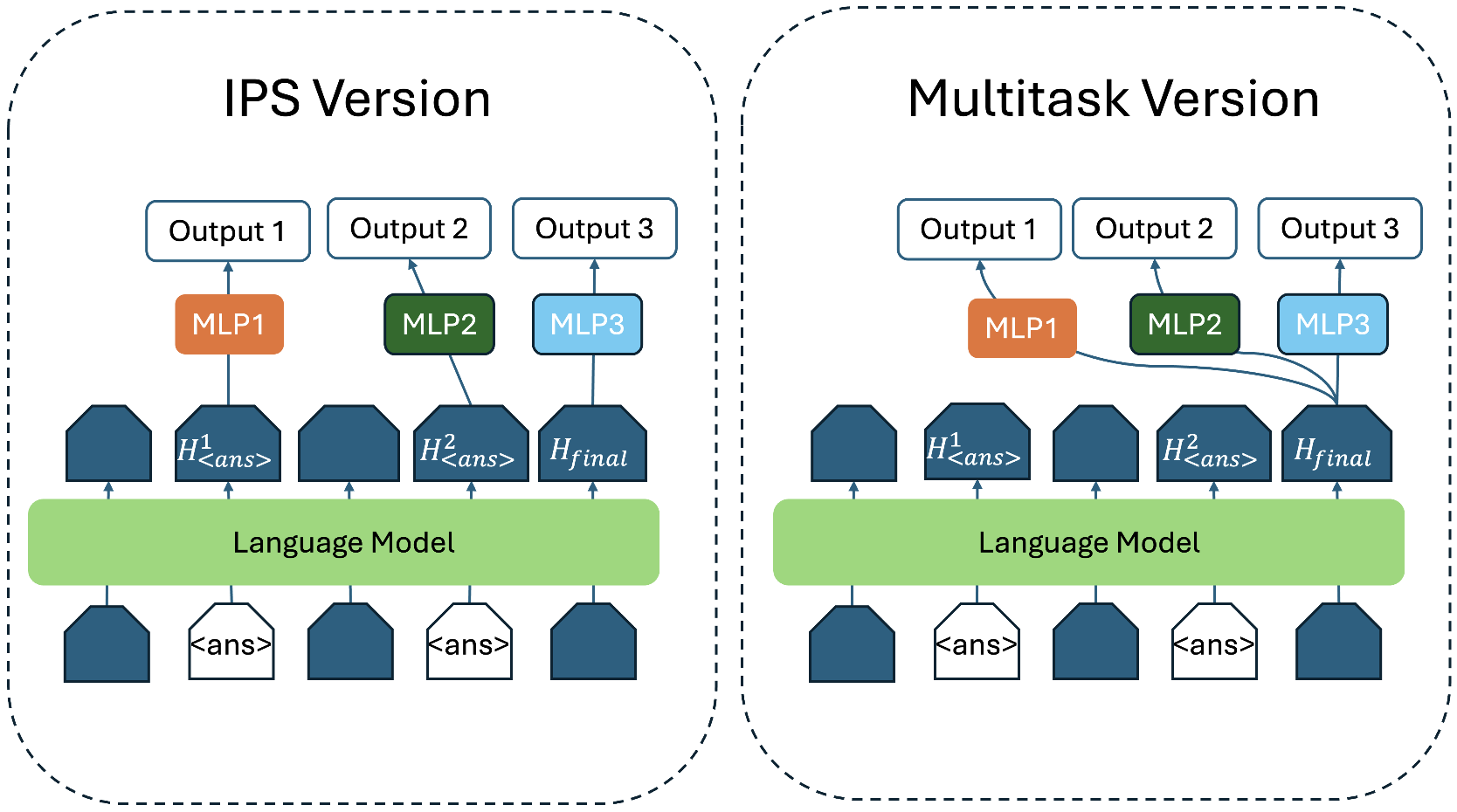}
    \caption{Difference between IPS and Multitask}
    \label{fig:IPS_Multitask}
\end{figure}

To further assess the impact of the \textit{CoT} reasoning structure in \textit{IPS}, we conducted an ablation study on the UCC task using a multitask variant. In this variant, a multi-head MLP is applied to the final token’s hidden state, with each head predicting labels for the four ancillary questions and the final question (\textbf{Figure~\ref{fig:IPS_Multitask}}). This contrasts with \textit{IPS}, which uses MLPs on hidden states from distinct positions in the sequence.
As shown in \textbf{Table~\ref{tab:sequential_qa_impact}}, while the multitask variant consistently outperforms the vanilla model, it does not match the performance of \textit{IPS}.

The experiments suggest that the performance gains of \textit{IPS} are attributable to two primary contributing factors: 
\begin{itemize}
    \item The incorporation of additional process supervision, quantified by the performance gap between the multitask and vanilla models $(\text{Multitask} - \text{Vanilla})$.

    \item The \textit{CoT} reasoning structure. Process supervision on the carefully placed \texttt{<ans>} tokens effectively leverages this additional information, as evidenced by the gap between \textit{IPS} and the multitask variant $(\text{IPS} - \text{Multitask})$.
\end{itemize}

\begin{table}[htbp]
\small
\space
\centering
\resizebox{\linewidth}{!}{\begin{tabular}{l | c  c  c}
    \hline
    \hline
     Models &    R@P60    &     R@P70    &    R@P80  \\ 
    % \hline
    % SigLIP  &   70.6 & 55.2 & 41.0  \\%\hline
    \hline
    LLaVA-OV-0.5B Vanilla &   71.6 & 63.1 & 52.1  \\%\hline
    LLaVA-OV-0.5B Multitask &  74.2 & 64.4 & 54.7 \\%\hline
    LLaVA-OV-0.5B \textit{IPS} &  \textbf{76.4} & \textbf{68.0} & \textbf{56.8}  \\%\hline
    \hline
    LLaVA-OV-7B Vanilla &  75.4 & 66.9 & 51.0 \\%\hline
    LLaVA-OV-7B Multitask &  75.6 & 66.1 & 52.5 \\%\hline
    LLaVA-OV-7B \textit{IPS} & \textbf{76.6} & \textbf{67.2} & \textbf{58.0} \\%\hline
    \hline
    \hline
    \end{tabular}
}
\caption{Performance (in \%) Comparison on UCC task: MLLMs in Vanilla, Multitask, and \textit{IPS} settings (LLaVA-OneVision 0.5B and 7B).}
\label{tab:sequential_qa_impact}
\end{table}

\begin{table*}[htbp]
\small
\centering
\begin{tabular}{l | c | c | c  c  c  c  c}
\hline
\hline
 Models &                   Data Size & F1   &P@R50 &R@P50 &R@P55 &R@P60 & R@P65  \\ 
\hline
LLaVA-OV-0.5b Vanilla &  \multirow{2}{*}{854k} & 48.6 & 46.5 & 46.7 & 41.8 & 37.1 & 29.6  \\%\hline
LLaVA-OV-0.5b \textit{IPS} &  & \textbf{53.0} & \textbf{55.2} & \textbf{54.9} & \textbf{50.2} & \textbf{45.2} & \textbf{40.3} \\%\hline
\hline
LLaVA-OV-0.5b Vanilla &  \multirow{2}{*}{1.7M} & 51.6 & 52.1 & 52.3 & 45.8 & 41.9 & 38.1  \\%\hline
LLaVA-OV-0.5b \textit{IPS} &  & \textbf{54.6} & \textbf{60.1} & \textbf{58.4} & \textbf{52.2} & \textbf{50.0} & \textbf{43.2} \\%\hline
\hline
LLaVA-OV-0.5b Vanilla &  \multirow{2}{*}{2.5M} & 53.2 & 55.4 & 54.8 & 51.0 & 46.5 & 41.0  \\%\hline
LLaVA-OV-0.5b \textit{IPS} &  & \textbf{55.6} & \textbf{60.4} & \textbf{60.0} & \textbf{55.4} & \textbf{51.0} & \textbf{43.0} \\%\hline
\hline
% LLaVA-OV-0.5b Vanilla &               & 54.8 & 60.2 & 58.2 & 53.8 & 50.0 & 42.8  \\%\hline
% LLaVA-OV-0.5b \textit{IPS} & 3.2m & 55.5 & 61.0 & 60.1 & 55.8 & 50.7 & 47.5 \\%\hline
% \hline
LLaVA-OV-0.5b Vanilla &  \multirow{2}{*}{4M} & 54.9 & 59.9 & 58.2 & 53.8 & 49.9 & 44.1  \\%\hline
LLaVA-OV-0.5b \textit{IPS} &   & \textbf{56.4} & \textbf{63.0} & \textbf{60.5} & \textbf{57.4} & \textbf{52.8} & \textbf{46.2} \\%\hline
\hline
\hline
\end{tabular}
\caption{\textit{IPS} with MLLM-generated process labels on ANSA dataset. Different sizes of training data are compared to demonstrate the consistency of this approach. (Performance displayed in \%)}
\label{tab:ANSA}
\end{table*}

\section{\textit{IPS}'s compatibility with autoregressive CoT training}

Chain-of-thought (CoT)–instructed fine-tuning is a widely adopted technique in multimodal reasoning (Ma et al., 2023\cite{ma2023adapting}). Compared with traditional CoT-centered fine-tuning, \textit{IPS} focuses on a different aspect of improvement. While CoT-centered training emphasizes letting the model auto-regressively generate CoT sequences before producing an answer, \textit{IPS} focuses on process-supervising a model within a fixed CoT-like input prompt.

In earlier experiments, we used randomly initialized, trainable MLP classifiers, which allowed us to compare the multi-task model with the \textit{IPS} model. In this experiment, each MLP is instead initialized with the built-in lm\_head from the original qwen2.5-vl checkpoint and kept frozen to ensure stability. The logits of selected tokens serve as the MLP outputs: for binary questions, the tokens are “No” and “Yes,” and for multiple-choice questions, the tokens are “A,” “B,” “C,” etc. This initialization preserves the model’s text generation ability, thereby preserving full compatibility with CoT generation.

\begin{table}[htbp]
\small
\space
\centering
\resizebox{\linewidth}{!}{\begin{tabular}{l | l | c  c  c}
    \hline
    \hline
     Task    &   Models &    R@P50   &     F1    \\ 
    \hline
   \multirow{4}{*}{ANSA-borderline}  & Qwen2.5-vl-7B Vanilla &  41.04 & 45.32   \\%\hline
   &   Qwen2.5-vl-7B CoT & 43.40 & 46.67  \\%\hline
   &   Qwen2.5-vl-7B \textit{IPS}  & 43.48 & 46.83  \\%\hline
    &  Qwen2.5-vl-7B CoT + \textit{IPS} & \textbf{45.38} & \textbf{47.66} \\%\hline
    \hline
    \end{tabular}
}
\caption{ANSA-borderline classification Performance (in \%) comparison of traditional CoT training, \textit{IPS}, and combined.}
\label{tab:cot_IPS_performance}
\end{table}

The ANSA-borderline dataset is a subset of the ANSA dataset. It contains ambiguous or borderline sexual content, providing a challenging testbed for reasoning-based classifiers. \textbf{Table~\ref{tab:cot_IPS_performance}} demonstrates that \textit{IPS} can be applied in parallel with traditional CoT. Notably, combining Ma et al.’s CoT method with \textit{IPS} achieves the latest state-of-the-art performance on the ANSA-borderline benchmark. This confirms that \textit{IPS} enhances CoT-style training without interfering with the model’s generative reasoning pipeline.

\section{Principles for Ancillary Question Generation}
\label{principle_for_ancillary}
Although generating ancillary questions require domain knowledge and could be different for different issue, several empirical principles remained consistent throughout this project. In practice, the key consideration is \textbf{coverage}, defined as the proportion of samples answering “Yes” to a given ancillary question. Questions with moderate coverage (20-80\% positive responses) provide informative supervision, whereas questions with coverage near 0\% or 100\% yield little to no informational gain. Furthermore, questions are preferred when their labels exhibit a strong correlation with the final target label.

\begin{figure}[ht]
    \centering  
    \includegraphics[width=5cm]{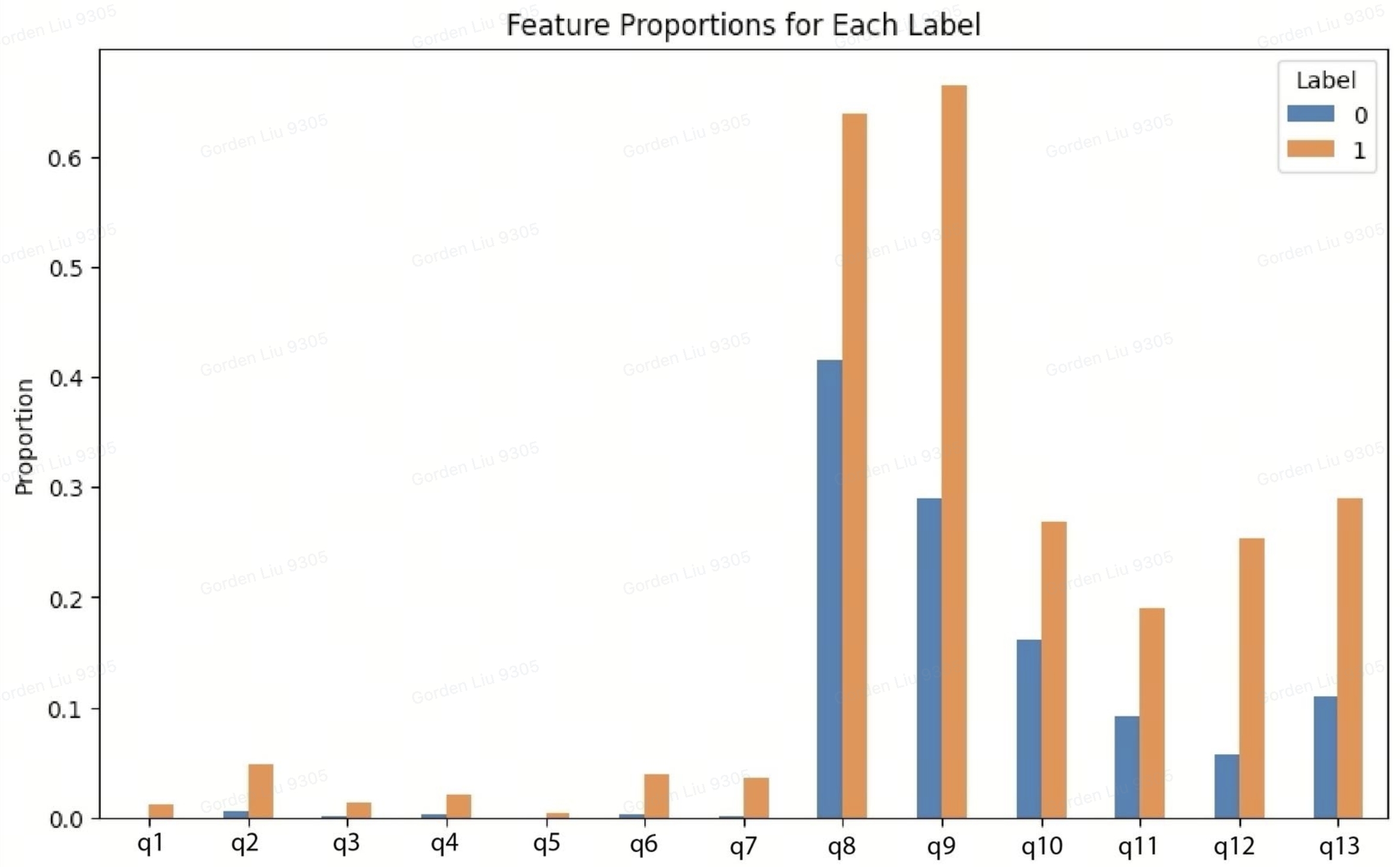}
    \caption{Positive Response Rate of Candidate Ancillary Questions. 13 candidate questions go through the first round of machine labeling in 1/5 of the dataset, evaluating it's coverage in non violation cases and violation cases respectively. Question example:"Is adult product mentioned in the content?". Details of the questions are omitted due to company privacy considerations.}
    \label{fig:ancillary-question}
\end{figure}

\begin{table}[htbp]
\small
\space
\centering
\resizebox{\linewidth}{!}{\begin{tabular}{l | l | c  c  c}
    \hline
    \hline
     Task    &   Models &  F1  &    R@P55   \\ 
    \hline
   \multirow{4}{*}{ANSA 1/5 training-set}  &  Vanilla &  48.6 & 41.8   \\%\hline
   &   \textit{IPS} with question 1-7 & 47.1 & 40.0  \\%\hline
   &   \textit{IPS} with question 1-2  & 47.5 & 40.9  \\%\hline
    &  \textit{IPS} with question 8-9 & \textbf{53.0} & \textbf{50.2} \\%\hline
    \hline
    \end{tabular}
}
\caption{Ablation (in \%) on ancillary-question coverage on the ANSA 1/5 training-set. Low-coverage questions (1--7) can hurt performance, while high-coverage questions (8--9) substantially improve over the vanilla baseline.}
\label{tab:ancillary-quick-eval}
\end{table}

During the development of the ANSA-IPS model, we designed 13 candidate ancillary questions and analyzed the distribution of ``Yes'' responses from the LLM annotator across negative (label 0) and positive (label 1) cases (\textbf{Figure~\ref{fig:ancillary-question}}). Early questions, though almost exclusively ``Yes'' in positive cases, appeared in fewer than 5\% of them, providing insufficient coverage. Later questions occurred in over 60\% of positive cases despite ``Yes'' responses in about 30\% of negative cases, offering a better balance between coverage and discriminative power. Using sparse-response questions initially led to performance below the baseline, whereas replacing them with higher-coverage questions (8 and 9) substantially improved results (\textbf{Table~\ref{tab:ancillary-quick-eval}}); these questions were ultimately used in \textbf{Table~\ref{tab:ANSA}}.

For UCC and the ANSA-borderline dataset, we relied on human-annotated results for the initial attempt and did not create our own set of candidate questions. Instead, we aligned directly with the questions already present in the annotation template used by human agents. For the HSD task, our first attempt was simple and effective; the prompt details are provided in Appendix~\ref{llm_prompt_hate}.

% For example, when iterating on the ANSA-IPS model, we initially developed 13 candidate ancillary questions and analyzed the distribution of “Yes” responses across negative cases (blue, label 0) and positive cases (orange, label 1) (Figure~\ref{fig:ancillary-question}). Although the first seven questions were almost exclusively answered “Yes” in positive cases, they occurred in fewer than 5\% of such cases, and thus provided insufficient coverage. In contrast, later questions exhibited broader coverage: despite producing “Yes” responses for approximately 30\% of negative cases, they occurred in over 60\% of positive cases, striking a more useful balance between coverage and discriminative power.

% When we iterate the first version of IPS model on ANSA, we first used questions from the sparse part, resulting in a model worse than it's vanilla comparison. Then we reduce the number of questions chosen(question 1 and 2) and use questions with better coverage (question 8 and 9), the results are shown in \textbf{Table~\ref{tab:ancillary-quick-eval}}. Questions with higher coverage (coverage 8 and 9) will significantly increase the model performance while question with sparse response will even slightly make the model perform worse. Question 8 and 9 are also the ancillary questions used in the \textbf{Table~\ref{tab:ANSA}}.

\section{Case Study on HSD Dataset}
\label{case_study}

From the case analysis in \textbf{Table~\ref{tab:case-comparison}}, \textit{IPS} demonstrates more accurate judgment for complex cases, such as ID=1268 and 2783. This indicates that \textit{IPS} facilitates the model's understanding of the policy by breaking down the problem.

Beyond aggregate accuracy, the per-dimension ancillary scores themselves serve as an interpretability trace that a free-text CoT output does not readily provide.
In case 2783, for example, the vanilla model emits only a single low-confidence score (0.015) and offers no way to audit why the prediction was made, whereas \textit{IPS} decomposes the decision into high ethnicity (0.980) and gender (0.926) signals with a near-zero religion signal (0.012), correctly surfacing the targeted group before producing the final label (0.991). In the clean case 1268, the low scores on all three ancillary dimensions give the moderator an immediate, structured justification for why the final prediction is safe.
This structured decomposition also has a deployment-scale benefit: because every ancillary prediction is a discrete classifier output, policy teams can directly monitor what fraction of the video ecosystem triggers each individual ancillary criterion over time, enabling targeted policy refinement that free-text reasoning chains cannot easily support.
\begin{table*}[htbp]
    \centering
    \renewcommand{\arraystretch}{0.9} 
    \resizebox{\textwidth}{!}{
    \begin{tabular}{c|p{4cm}|c|c|c|c|c|c|c}
        \hline
        \hline
        id & text & image & label & Vanilla score & Ethnicity score& Gender score& Religion score& \textit{IPS}-final score \\
        \hline
        
        2783 & a picture of my great great great grand grandfather's tractors they don't make them like they used to & \includegraphics[width=5cm]{6.pdf}& 1 & 0.015 &0.980&0.926&0.012& 0.991 \\
        \hline
        % 9467 & when mom asks where all the jews went & \includegraphics[width=5cm]{9.pdf}& 1 & 0.034 &0.998&0.996&0.959& 0.986 \\
        % \hline
        
        1268 & father and daughter bonding time is the best like and share if you agree & \includegraphics[width=5cm]{5.pdf}& 0 & 0.889 &0.114&0.021&0.048& 0.001 \\
        \hline
        
        4538 & bows here, bows there, bows everywhere & \includegraphics[width=5cm]{8.pdf}& 0 & 0.843 &0.055&0.0168&0.006& 0.009 \\
        \hline
        \hline
    \end{tabular}
    }
    \caption{Comparison of Vanilla and \textit{IPS} Scores on HSD Cases in MM-Soc Validation Set. \textit{IPS} achieves more accurate predictions for complex problems by decomposing the issue.}
    \label{tab:case-comparison}
\end{table*}

\section{Model Training Details}
\label{model_training}
For MLLM, we use the LLaVA-OneVision model \cite{li2024llava} as the VL backbone, selecting its 0.5B and 7B versions for their balance between computational efficiency and strong multimodal reasoning capabilities.
For SFT, we adopt LoRA~\cite{hu2021lora} with default parameters ($rank = 128, alpha = 256$) to optimize training efficiency.
% [TO WRITE: GPU, training]
We trained the model with 16 A100-SXM-80GB GPU. The typical 5-epoch fine-tuning and evaluation process on UCC dataset takes 18 hours for 7b version and 10 hours for the 0.5b version. % The fine-tuning achieves optimal performance after three epochs of training. 

To process video input, same number of frames are sampled from each video. All frames representations are converted to the 3d-Array format and uniformly resized to the same size.

After the fixed sized visual tokens, we designed a fixed m-token prompt for \textit{IPS}, which is appended to the end of language input, facilitating structured process supervision. The text token from data sample is clipped to (256-m) token in order to avoid token overflow.

\section{Limitations}

Our current evaluation focuses on domain-specific classification tasks, and we do not assess the text generation capabilities of the fine-tuned model. Exploring the generative potential of \textit{IPS}-based models remains an avenue for future work.

While ancillary questions enhance the model’s reasoning toward the final decision, designing effective ones requires domain expertise, as they must align closely with the underlying task. Poorly formulated or irrelevant questions may fail to provide useful guidance and could even impair performance. In addition, scaling this method to datasets without human-annotated process labels requires ancillary questions to be interpretable for general-purpose MLLMs. Empirical guidelines for constructing high-quality ancillary questions are provided in Appendix~\ref{principle_for_ancillary}.

Although \textit{IPS} is designed to be noise-aware and tolerant of inaccuracies in ancillary labels, accurate final labels are still essential to ensure stable training and reliable evaluation.

\section{Prompt for UCC dataset with MLLM Process Annotation}
\label{llm_prompt}
The prompt consists of two parts:
\begin{enumerate}
    \item Image-based Prompt: A list of image collections.
    \item Text-based Prompt: A sequence of text-based questions.
\end{enumerate}

\subsection{Image-based Prompt}
The image collection list includes 1 to 16 images, which are encoded using b64encode and then concatenated after the questions.

\subsection{Text-based Prompt}

\textbf{1. Watermark presence.} 
"'Watermark' is like '@username' from social media, not simple timestamp. Each image is considered as one image. Count the number of images with watermarks in the album."

\textbf{2. Whether it is UGC (User-Generated Content).} 
"UGC (User Generated Content) is considered as content is generated by regular users, such as selfies, artistic creations, life recordings, or concatenated images from online sources combined with self-created content. The opposite of UGC is PGC (Professionally Generated Content). PGC refers to content such as pictures of celebrities in entertainment/sports/politics, screenshots, posters, or coverage from TV series, movies, documentaries, and other platforms. Each image is considered as one image. Count the number of UGC images in the album."

\textbf{3. Whether the image and the text title are relevant.} 
"Original text is defined as content with emotional words (e.g., 'good,' 'happy,' 'disgusting') or symbols, subjective comments (e.g., 'I think the Doors are the best rock band'), or narrative storytelling (e.g., 'This movie tells the story of...'). Simple expressions without detail, like song lyrics or standalone sentences, are considered non-original. Determine if the given text is original."

% \textbf{4. Whether the image and the text content are relevant.} 
% "Original text is defined as content with emotional words (e.g., 'good,' 'happy,' 'disgusting') or symbols, subjective comments (e.g., 'I think the Doors are the best rock band'), or narrative storytelling (e.g., 'This movie tells the story of...'). Simple expressions without detail, like song lyrics or standalone sentences, are considered non-original. Determine if the given text is original."

\textbf{4. Whether the image and the overall theme of the image collection are relevant.}
"Each image is considered as one image. Count the number of images whose content is related to the overall theme of the album."

\section{Prompt for MM-Soc Hate-speech Detection with MLLM Process Annotation}
\label{llm_prompt_hate}
The prompt consists of two parts:
\begin{enumerate}
    \item Image-based Prompt: one meme image.
    \item Text-based Prompt: a sequence of three text-based questions answered within a single session.
\end{enumerate}

\subsection{Image-based Prompt}
The image is encoded using \texttt{b64encode} and concatenated after the questions.

\subsection{Text-based Prompt}
The three ancillary questions are presented together in one session:

\textbf{1. Ethnicity or Country.}
``Does the image and the given text contain satirical, discriminatory, harmful, cursing, racial, or other hateful content toward a certain ethnicity or country?''

\textbf{2. Gender or a Certain Group of People.}
``Does the image and the given text contain satirical, discriminatory, harmful, cursing, racial, or other hateful content toward a certain gender or a certain group of people?''

\textbf{3. Religion.}
``Does the image and the given text contain satirical, discriminatory, harmful, cursing, racial, or other hateful content toward a certain religion?''

Each question is answered independently with a binary response in \{0, 1\}, where 0 indicates no hateful content and 1 indicates hateful content in the targeted dimension.
\end{document}